\pdfoutput=1
\documentclass[11pt]{article}
\usepackage{emnlp-ijcnlp-2019}
\usepackage{times}
\usepackage{latexsym}
\usepackage{url}
\usepackage{booktabs}
\usepackage[shortlabels]{enumitem}
\usepackage{graphicx}

\usepackage{footnote}
\makesavenoteenv{tabular}
\makesavenoteenv{table}

\aclfinalcopy 

\definecolor{DarkRed}{RGB}{130,25,0}

\title{ner and pos when nothing is capitalized}

\author{Stephen Mayhew, Tatiana Tsygankova, Dan Roth\\
	    University of Pennsylvania\\
	    Philadelphia, PA, 19104\\
	    {\tt \{mayhew, ttasya, danroth\}@seas.upenn.edu}}
\date{}

\begin{document}

\maketitle

\begin{abstract}

For those languages which use it, capitalization is an important signal for the fundamental NLP tasks of Named Entity Recognition (NER) and Part of Speech (POS) tagging. In fact, it is such a strong signal that model performance on these tasks drops sharply in common lowercased scenarios, such as noisy web text or machine translation outputs. In this work, we perform a systematic analysis of solutions to this problem in English, modifying only the casing of the train or test data using lowercasing and truecasing methods. While prior work and first impressions might suggest training a caseless model, or using a truecaser at test time, we show that the most effective strategy is a concatenation of cased and lowercased training data, producing a single model with high performance on both cased and uncased text. As shown in our experiments, this result holds across tasks and input representations. Finally, we show that our proposed solution gives an 8\% F1 improvement in mention detection on noisy out-of-domain Twitter data.

\end{abstract}

\section{Introduction}
Many languages use capitalization in text, often to indicate named entities. For tasks that are concerned with named entities, such as \textit{named entity recognition} (NER) and \textit{part of speech tagging} (POS), this is an important signal, and models for these tasks nearly always retain it in training.\footnote{For POS tagging, this happens in tagsets that explicitly mark proper nouns, such as the Penn Treebank tagset.}

\begin{table}[]
\small
    \centering
    \begin{tabular}{llrr}
    \toprule
        & & \multicolumn{2}{c}{Test} \\
        \cmidrule{3-4}
        Tool & Task & Cased & Uncased \\ 
         \midrule
        BiLSTM-CRF w/ ELMo & NER & 92.45 & 34.46 \\ 
        \midrule
        BiLSTM-CRF w/ ELMo & POS & 97.85 & 88.66 \\
         \bottomrule
    \end{tabular}
    \caption{Modern tools trained on cased data perform well on cased test data, but poorly on uncased (lowercased) test data. For NER, we evaluate on the testb set of CoNLL 2003, and the scores are reported as F1. For POS, we evaluate on PTB sections 22-24, and the scores represent accuracy. ELMo refers to contextual representations from \citet{PNIGCLZ18}. }
    \label{tab:motivation}
\end{table}

But capitalization is not always available. For example, informal user-generated texts can have inconsistent capitalization, and similarly the outputs of speech recognition or machine translation are traditionally without case. Ideally we would like a model to perform equally well on both cased and uncased text, in contrast with current models. Table \ref{tab:motivation} demonstrates how popular modern systems trained on cased data perform well on cased data, but suffer dramatic performance drops when evaluated on lowercased text. 

Prior solutions have included models trained on lowercase text, or models that automatically recover capitalization from lowercase text, known as \textit{truecasing}. There has a been a substantial body of literature on the effect of truecasing applied after speech recognition \cite{GravanoJaBa09}, machine translation \cite{WangKnMa06}, or social media \cite{NebhiBoGo15}. A few works that evaluate on downstream tasks (including NER and POS) show that truecasing improves performance, but they do not demonstrate that truecasing is \textit{the best way} to improve performance. 

In this paper, we evaluate two foundational NLP tasks, NER and POS, on cased text and lowercased text, with the goal of maximizing the average score regardless of casing. To achieve this goal, we explore a number of simple options that consist of modifying the casing of the train or test data. Ultimately we propose a simple preprocessing method for training data that results in a single model with high performance on both cased and uncased datasets.

\section{Related Work}
\label{sec:relatedwork}
This problem of robustness in casing has been studied in the context of NER and truecasing.

\vspace{0.1cm}
\textbf{Robustness in NER}~~
A practical, common solution to this problem is summarized by the Stanford CoreNLP system \cite{corenlp}: train on uncased text, or use a truecaser on test data.\footnote{\url{https://stanfordnlp.github.io/CoreNLP/caseless.html}} We include these suggested solutions in our analysis below.

In one of the few works that address this problem directly, \citet{chieu2002teaching} describe a method similar to co-training for training an upper case NER, in which the predictions of a cased system are used to adjudicate and improve those of an uncased system. One difference from ours is that we are interested in having a single model that works on upper or lowercased text. When tagging text in the wild, one cannot know \textit{a priori} if it is consistently cased or not. 

\vspace{0.1cm}
\textbf{Truecasing}~~
Truecasing presents a natural solution for situations with noisy or uncertain text capitalization. It has been studied in the context of many fields, including speech recognition \cite{BrownCo01,GravanoJaBa09}, and machine translation \cite{WangKnMa06}, as the outputs of these tasks are traditionally lowercased.

\citet{LIRK03} proposed a statistical, word-level, language-modeling based method for truecasing, and experimented on several downstream tasks, including NER. \citet{NebhiBoGo15} examine truecasing in tweets using a language model method and evaluate on both NER and POS.

More recently, a neural model for truecasing has been proposed by \citet{SusantoChLu16}, in which each character is associated with a label \textit{U} or \textit{L}, for upper and lower case respectively. This neural character-based method outperforms word-level language model-based prior work.

\begin{table}

    \centering
    \begin{tabular}{llr}
    \toprule
        System & Test set & F1 \\
    \midrule
    \cite{SusantoChLu16} & Wikipedia & 93.19 \\
    \midrule
    BiLSTM & Wikipedia & 93.01 \\
        & CoNLL Train & 78.85 \\
        & CoNLL Test & 77.35 \\
        & PTB 01-18 & 86.91 \\
        & PTB 22-24 & 86.22 \\
    \bottomrule
    \end{tabular}
    \caption{Truecaser word-level performance on English data. This truecaser is trained on the Wikipedia corpus.  \textit{Wikipedia} refers to the test set from \citet{CosterKa11}. \textit{CoNLL Test} refers to \textit{testb}. \textit{PTB} is the Penn Treebank.}
    \label{tab:truecase_perf}
\end{table}

\section{Truecasing Experiments}
\label{sec:truecasing}
 We use our own implementation of the neural method described in \citet{SusantoChLu16} as the truecaser used in our experiments.\footnote{\url{cogcomp.org/page/publication_view/881}} Briefly, each sentence is split into characters (including spaces) and modeled with a 2-layer bidirectional LSTM, with a linear binary classification layer on top.
 
 We train the truecaser on a dataset from Wikipedia, originally created for text simplification \cite{CosterKa11}, but commonly used for evaluation in truecasing papers \cite{SusantoChLu16}. This task has the convenient property that if the data is well-formed, then supervision is free. We evaluate this truecaser on several data sets, measuring F1 on the word level (see Table \ref{tab:truecase_perf}). At test time, all text is lowercased, and case labels are predicted.

First, we evaluate the truecaser on the same test set as \citet{SusantoChLu16} in order to show that our implementation is near to the original. Next, we measure truecasing performance on plain text extracted from the CoNLL 2003 English \cite{TjongDe03} and Penn Treebank \cite{MarcusSaMa03} train and test sets. These results contain two types of errors: idiosyncratic casing in the gold data and failures of the truecaser. However, from the high scores in the Wikipedia experiment, we suppose that much of the score drop comes from idiosyncratic casing. This point is important: if a dataset contains idiosyncratic casing, then it is likely that NER or POS models have fit to that casing (especially with these two wildly popular datasets). As a result, truecasing, since it can't recover these idiosyncrasies, is not likely to be the best plan. 

Notably, the scores on CoNLL are especially low, likely because of elements such as titles, bylines, and documents that contain league standings and other sports results written in uppercase.

The higher scores on Penn Treebank corpus suggest that the capitalization standards are more traditional. Many errors are where the truecaser fails to correctly capitalize such words as ``Federal" or ``Central". In addition, there are many occasions where the truecaser fails to capitalize named entities, for example ``Mr. susulu".

\section{Methods}
\label{sec:methods}
In this section, we introduce our proposed solutions. In all experiments, we constrain ourselves to only change the casing of the training or testing data with no changes to the architectures of the models in question. This isolates the importance of dealing with casing, and makes our observations applicable to situations where modifying the model is not feasible, but retraining is possible. 

Our experiments aim to answer the extremely common situation in which capitalization is noisy or inconsistent (as with inputs from the internet). In light of this goal, we evaluate each experiment on both cased and lowercased test data, reporting individual scores as well as the average. Our experiments on lowercase text can also give insight on best practices for when test data is known to be all lowercased (as with the outputs of some upstream system).

We experiment on five different data casing scenarios described below.

\begin{enumerate}
    \item \textbf{Train on cased}~~
Simply apply a model trained on cased data to unmodified test data, as in Table \ref{tab:motivation}.
    
    \item \textbf{Train on uncased}~~
Lowercase the training data and retrain. At test time, we lowercase all test data. If we did not do this, then scores on the cased test set would suffer because of casing mismatch between train and test. Since lowercasing costs nothing, we can improve average scores this way. As such, cased and uncased test data will have the same score. 

    \item \textbf{Train on cased+uncased}~~
Concatenate original cased and lowercased training data and retrain a model. Test data is unmodified.

Since this concatenation results in twice the number of training examples than other methods, we also experimented with randomly lowercasing 50\% of the sentences in the original training corpus. We refer to this experiment as \textit{3.5 Half Mixed}. We also tried ratios of 40\% and 60\%, but these were slightly worse than 50\% in our evaluations.

\item \textbf{Train on cased, test on truecased}~~
Do nothing to the train data, but truecase the test data. Since we lowercase text before truecasing it, the cased and uncased test data will have the same score.

\item \textbf{Truecase train and test}~~
Truecase the train data and retrain. Truecase the test data also. As in experiment 4, cased and uncased test data will have the same score.
\end{enumerate}

One way to look at these experiments is as dropout for capitalization, where a sentence is lowercased with respect to the original with probability $p \in [0,1]$. In experiment 1, $p = 0$. In experiment 2, $p = 1$. In experiment 3, $p = 0.5$. Our implementation is somewhat different from standard dropout in that our method is a preprocessing step, not done randomly at each epoch.

\section{Experiments}
Before we show results, we will describe our experimental setup. We emphasize that our goal is to experiment with strong models in noisy settings, not to obtain state-of-the-art scores on any dataset.

\subsection{NER}
We use the standard BiLSTM-CRF architecture for NER \cite{MaHo16}, using an AllenNLP implementation \cite{GGNTDLPSZ18}. 

We experiment with pre-trained contextual embeddings, ELMo \cite{PNIGCLZ18}, which are generated for each word in a sentence, and concatenated with GloVe word vectors (lowercased) \cite{PenningtonSoMa14}, and character embeddings. ELMo embeddings are trained with cased inputs, meaning that there will be some mismatch when generating embeddings for uncased text. 

In all experiments, we train on English CoNLL 2003 Train data \cite{TjongDe03} and evaluate on the CoNLL 2003 Test data (testb). We always evaluate on two different versions: the original version, and a version with all casing removed (e.g. everything lowercase).

\subsection{POS Tagging}
We use a neural POS tagging model built with a BiLSTM-CRF \cite{MaHo16}, and GloVe embeddings \cite{PenningtonSoMa14}, character embeddings, and ELMo pre-trained contextual embeddings \cite{PNIGCLZ18}. 

As our experimental data, we use the Penn Treebank \cite{MarcusSaMa03}, and follow the training splits of \cite{LDBTFAML15}, namely 01-18 for train, 19-21 for validation, 22-24 for testing. As with NER, we evaluate on both original and lowercased versions of test.

\begin{table}
    \centering
    \begin{tabular}{lrrr}
    \toprule
   Exp. & Test (C) & Test (U) & Avg \\
    \midrule
         1. Cased & \textbf{92.45} & 34.46 & 63.46 \\
         2. Uncased & 89.32 & 89.32 & 89.32 \\
         3. C+U & 91.67 & 89.31 & \textbf{90.49} \\
         ~~3.5. Half Mixed & 91.68 & 89.05 & 90.37 \\
         4. Truecase Test & 82.93 & 82.93 & 82.93 \\
         5. Truecase All & 90.25 & \textbf{90.25} & 90.25 \\
    \bottomrule
    \end{tabular}
    \caption{Results from NER+ELMo experiments, tested on CoNLL 2003 English test set. \textit{C} and \textit{U} are Cased and Uncased respectively. All scores are F1.}
    \label{tab:NER}
\end{table}

\section{Results}

Results for NER are shown in Table \ref{tab:NER}, and results for POS are shown in Table \ref{tab:POS}. There are several interesting observations to be made. 

Primarily, our experiments show that the approach with the most promising results was experiment 3: training on the concatenation of original and lowercased data. Lest one might think this is because of the double-size training corpus, results from experiment 3.5 are either in second place (for NER) or slightly ahead (for POS).

Conversely, we show that the folk-wisdom approach of truecasing the test data (experiment 4) does not perform well. The underwhelming performance can be explained by the mismatch in casing standards as seen in Section \ref{sec:truecasing}.  However, experiment 5 shows that if the training data is also truecased, then the performance is good, especially in situations where the test data is known to contain no case information.

Training only on uncased data gives good performance in both NER and POS -- in fact the highest performance on uncased text in POS -- but never reaches the overall average scores from experiment 3 or 3.5.

We have repeated these experiments for NER in several different settings, including using only static embeddings, using a non-neural truecaser, and using BERT uncased embeddings \cite{DCLT19}. While the relative performance of the experiments varied, the conclusion was the same: training on cased and uncased data produces the best results. 

When using uncased BERT embeddings, we found that performance on the uncased test set (U) was typically higher than that of ElMo, while the maximum performance on the cased test set (C) was typically lower. This again exemplifies the challenge of using capitalization as a signal while being robust to its absence.

\begin{table}
    \centering
    \begin{tabular}{lrrr}
    \toprule
   Exp. & Test (C) & Test (U) & Avg \\
    \midrule
    1. Cased & \textbf{97.85} & 88.66 & 93.26 \\
    2. Uncased & 97.45 & \textbf{97.45} & 97.45 \\
    3. C+U & 97.79 & 97.35 & 97.57 \\
    ~~3.5. Half Mixed & \textbf{97.85} & 97.36 & \textbf{97.61} \\
    4. Truecase Test & 95.21 & 95.21 & 95.21 \\
    5. Truecase All & 97.38 & 97.38 & 97.38 \\
    \bottomrule
    \end{tabular}
    \caption{Results from POS+ELMo experiments, tested on WSJ 22-24, from PTB. \textit{C} and \textit{U} are Cased and Uncased respectively. All scores are accuracies.}
    \label{tab:POS}
\end{table}

\section{Application: Improving NER Performance on Twitter}
To further test our results, we look at the Broad Twitter Corpus\footnote{\url{https://github.com/GateNLP/broad\_twitter\_corpus}} \cite{DerczynskiBoRo16}, a dataset comprised of tweets gathered from a broad variety of genres, and including many noisy and informal examples. Since we are testing the robustness of our approach, we use a model trained on CoNLL 2003 data. Naturally, in any cross-domain experiment, one will obtain higher scores by training on in-domain data. However, our goal is to show that our methods produce a more robust model on out-of-domain data, not to maximize performance on this test set.  We use the recommended test split of section F, containing 3580 tweets of varying length and capitalization quality. 

Since the train and test corpora are from different domains, we evaluate on the level of mention detection, in which all entity types are collapsed into one. The Broad Twitter Corpus has no annotations for MISC types, so before converting to a single generic type, we remove all MISC predictions from our model. 

Results are shown in Table \ref{tab:BTC}, and a familiar pattern emerges. Experiment 3 outperforms experiment 1 by 8 points F1, followed by experiment 3.5 and experiment 5, showing that our approach holds when evaluated on a real-world data set.

\begin{table}
    \centering
    \begin{tabular}{lr}
    \toprule
   Exp. & Mention Detection F1\\
    \midrule
    1. Cased &  58.63 \\
    2. Uncased & 53.13 \\
    3. C+U & \textbf{66.14} \\
    ~~3.5. Half Mixed & 64.69 \\
    4. Truecase Test & 58.22 \\
    5. Truecase All & 62.66 \\
    \bottomrule
    \end{tabular}
    \caption{Results on NER+ELMo on the Broad Twitter Corpus, set F, measured as mention detection F1.}
    \label{tab:BTC}
\end{table}

\section{Conclusion}

We have performed a systematic analysis of the problem of unknown casing in test data for NER and POS models. We show that commonly-held suggestions (namely, lowercase train and test data, or truecase test data) are rarely the best. Rather, the most effective strategy is a concatenation of cased and lowercased training data. We have demonstrated this with experiments in both NER and POS, and have further shown that the results play out in real-world noisy data.

\section{Acknowledgments}
For their valuable feedback and suggestions, we would like to thank Jordan Kodner, Shyam Upadhyay, and Nitish Gupta.

This work was supported by Contracts HR0011-15-C-0113 and HR0011-18-2-0052 with the US Defense Advanced Research Projects Agency (DARPA). Approved for Public Release, Distribution Unlimited. The views expressed are those of the authors and do not reflect the official policy or position of the Department of Defense or the U.S. Government.

\bibliography{mybib,ccg,cited}

\begin{thebibliography}{18}
\expandafter\ifx\csname natexlab\endcsname\relax\def\natexlab#1{#1}\fi

\bibitem[{Brown and Coden(2001)}]{BrownCo01}
Eric~W Brown and Anni~R Coden. 2001.
\newblock Capitalization recovery for text.
\newblock In \emph{Workshop on Information Retrieval Techniques for Speech
  Applications}, pages 11--22. Springer.

\bibitem[{Chieu and Ng(2002)}]{chieu2002teaching}
Hai~Leong Chieu and Hwee~Tou Ng. 2002.
\newblock Teaching a weaker classifier: Named entity recognition on upper case
  text.
\newblock In \emph{Proceedings of the 40th Annual Meeting on Association for
  Computational Linguistics}, pages 481--488. Association for Computational
  Linguistics.

\bibitem[{Coster and Kauchak(2011)}]{CosterKa11}
Will Coster and David Kauchak. 2011.
\newblock \href {https://www.aclweb.org/anthology/W11-1601} {Learning to
  simplify sentences using {W}ikipedia}.
\newblock In \emph{Proceedings of the Workshop on Monolingual Text-To-Text
  Generation}, pages 1--9, Portland, Oregon. Association for Computational
  Linguistics.

\bibitem[{Derczynski et~al.(2016)Derczynski, Bontcheva, and
  Roberts}]{DerczynskiBoRo16}
Leon Derczynski, Kalina Bontcheva, and Ian Roberts. 2016.
\newblock \href {https://www.aclweb.org/anthology/C16-1111} {Broad twitter
  corpus: A diverse named entity recognition resource}.
\newblock In \emph{Proceedings of {COLING} 2016, the 26th International
  Conference on Computational Linguistics: Technical Papers}, pages 1169--1179,
  Osaka, Japan. The COLING 2016 Organizing Committee.

\bibitem[{Devlin et~al.(2019)Devlin, Chang, Lee, and Toutanova}]{DCLT19}
Jacob Devlin, Ming-Wei Chang, Kenton Lee, and Kristina Toutanova. 2019.
\newblock \href {https://doi.org/10.18653/v1/N19-1423} {{BERT}: Pre-training of
  deep bidirectional transformers for language understanding}.
\newblock In \emph{Proceedings of the 2019 Conference of the North {A}merican
  Chapter of the Association for Computational Linguistics: Human Language
  Technologies, Volume 1 (Long and Short Papers)}, pages 4171--4186,
  Minneapolis, Minnesota. Association for Computational Linguistics.

\bibitem[{Gardner et~al.(2018)Gardner, Grus, Neumann, Tafjord, Dasigi, Liu,
  Peters, Schmitz, and Zettlemoyer}]{GGNTDLPSZ18}
Matt Gardner, Joel Grus, Mark Neumann, Oyvind Tafjord, Pradeep Dasigi,
  Nelson~F. Liu, Matthew Peters, Michael Schmitz, and Luke Zettlemoyer. 2018.
\newblock \href {https://doi.org/10.18653/v1/W18-2501} {{A}llen{NLP}: A deep
  semantic natural language processing platform}.
\newblock In \emph{Proceedings of Workshop for {NLP} Open Source Software
  ({NLP}-{OSS})}, pages 1--6, Melbourne, Australia. Association for
  Computational Linguistics.

\bibitem[{Gravano et~al.(2009)Gravano, Jansche, and Bacchiani}]{GravanoJaBa09}
Agustin Gravano, Martin Jansche, and Michiel Bacchiani. 2009.
\newblock Restoring punctuation and capitalization in transcribed speech.
\newblock In \emph{Acoustics, Speech and Signal Processing, 2009. ICASSP 2009.
  IEEE International Conference on}, pages 4741--4744. IEEE.

\bibitem[{Ling et~al.(2015)Ling, Dyer, Black, Trancoso, Fermandez, Amir,
  Marujo, and Lu{\'\i}s}]{LDBTFAML15}
Wang Ling, Chris Dyer, Alan~W Black, Isabel Trancoso, Ram{\'o}n Fermandez,
  Silvio Amir, Lu{\'\i}s Marujo, and Tiago Lu{\'\i}s. 2015.
\newblock \href {https://doi.org/10.18653/v1/D15-1176} {Finding function in
  form: Compositional character models for open vocabulary word
  representation}.
\newblock In \emph{Proceedings of the 2015 Conference on Empirical Methods in
  Natural Language Processing}, pages 1520--1530, Lisbon, Portugal. Association
  for Computational Linguistics.

\bibitem[{Lita et~al.(2003)Lita, Ittycheriah, Roukos, and Kambhatla}]{LIRK03}
Lucian~Vlad Lita, Abe Ittycheriah, Salim Roukos, and Nanda Kambhatla. 2003.
\newblock {tRuEasIng}.
\newblock In \emph{Proceedings of the 41st Annual Meeting on Association for
  Computational Linguistics-Volume 1}, pages 152--159. Association for
  Computational Linguistics.

\bibitem[{Ma and Hovy(2016)}]{MaHo16}
Xuezhe Ma and Eduard Hovy. 2016.
\newblock \href {https://doi.org/10.18653/v1/P16-1101} {End-to-end sequence
  labeling via bi-directional {LSTM}-{CNN}s-{CRF}}.
\newblock In \emph{Proceedings of the 54th Annual Meeting of the Association
  for Computational Linguistics (Volume 1: Long Papers)}, pages 1064--1074,
  Berlin, Germany. Association for Computational Linguistics.

\bibitem[{Manning et~al.(2014)Manning, Surdeanu, Bauer, Finkel, Bethard, and
  McClosky}]{corenlp}
Christopher~D. Manning, Mihai Surdeanu, John Bauer, Jenny Finkel, Steven~J.
  Bethard, and David McClosky. 2014.
\newblock \href {http://www.aclweb.org/anthology/P/P14/P14-5010} {The
  {Stanford} {CoreNLP} natural language processing toolkit}.
\newblock In \emph{Association for Computational Linguistics (ACL) System
  Demonstrations}, pages 55--60.

\bibitem[{Marcus et~al.(1993)Marcus, Santorini, and
  Marcinkiewicz}]{MarcusSaMa03}
Mitchell~P. Marcus, Beatrice Santorini, and Mary~Ann Marcinkiewicz. 1993.
\newblock \href {https://www.aclweb.org/anthology/J93-2004} {Building a large
  annotated corpus of {E}nglish: The {P}enn {T}reebank}.
\newblock \emph{Computational Linguistics}, 19(2):313--330.

\bibitem[{Nebhi et~al.(2015)Nebhi, Bontcheva, and Gorrell}]{NebhiBoGo15}
Kamel Nebhi, Kalina Bontcheva, and Genevieve Gorrell. 2015.
\newblock Restoring capitalization in\# tweets.
\newblock In \emph{Proceedings of the 24th International Conference on World
  Wide Web}, pages 1111--1115. ACM.

\bibitem[{Pennington et~al.(2014)Pennington, Socher, and
  Manning}]{PenningtonSoMa14}
Jeffrey Pennington, Richard Socher, and Christopher Manning. 2014.
\newblock \href {https://doi.org/10.3115/v1/D14-1162} {{G}love: Global vectors
  for word representation}.
\newblock In \emph{Proceedings of the 2014 Conference on Empirical Methods in
  Natural Language Processing ({EMNLP})}, pages 1532--1543, Doha, Qatar.
  Association for Computational Linguistics.

\bibitem[{Peters et~al.(2018)Peters, Neumann, Iyyer, Gardner, Clark, Lee, and
  Zettlemoyer}]{PNIGCLZ18}
Matthew Peters, Mark Neumann, Mohit Iyyer, Matt Gardner, Christopher Clark,
  Kenton Lee, and Luke Zettlemoyer. 2018.
\newblock \href {https://doi.org/10.18653/v1/N18-1202} {Deep contextualized
  word representations}.
\newblock In \emph{Proceedings of the 2018 Conference of the North {A}merican
  Chapter of the Association for Computational Linguistics: Human Language
  Technologies, Volume 1 (Long Papers)}, pages 2227--2237, New Orleans,
  Louisiana. Association for Computational Linguistics.

\bibitem[{Susanto et~al.(2016)Susanto, Chieu, and Lu}]{SusantoChLu16}
Raymond~Hendy Susanto, Hai~Leong Chieu, and Wei Lu. 2016.
\newblock \href {https://doi.org/10.18653/v1/D16-1225} {Learning to capitalize
  with character-level recurrent neural networks: An empirical study}.
\newblock In \emph{Proceedings of the 2016 Conference on Empirical Methods in
  Natural Language Processing}, pages 2090--2095, Austin, Texas. Association
  for Computational Linguistics.

\bibitem[{Tjong Kim~Sang and De~Meulder(2003)}]{TjongDe03}
Erik~F Tjong Kim~Sang and Fien De~Meulder. 2003.
\newblock \href {https://aclweb.org/anthology/W/W03/W03-0419.pdf} {Introduction
  to the conll-2003 shared task: Language-independent named entity
  recognition}.
\newblock In \emph{Proc. of the Annual Meeting of the North American
  Association of Computational Linguistics (NAACL)}.

\bibitem[{Wang et~al.(2006)Wang, Knight, and Marcu}]{WangKnMa06}
Wei Wang, Kevin Knight, and Daniel Marcu. 2006.
\newblock \href {https://www.aclweb.org/anthology/N06-1001} {Capitalizing
  machine translation}.
\newblock In \emph{Proceedings of the Human Language Technology Conference of
  the {NAACL}, Main Conference}, pages 1--8, New York City, USA. Association
  for Computational Linguistics.

\end{thebibliography}
\bibliographystyle{acl_natbib}

\end{document}